\documentclass[10pt, a4paper]{article}
\usepackage{lrec2026}
\usepackage{multibib}
\newcites{languageresource}{Language Resources}
\usepackage{graphicx}
\usepackage{tabularx}
\usepackage{soul}
\usepackage{hyperref}
\usepackage{booktabs}
\usepackage{xcolor}
\usepackage{multirow}
\usepackage{amsmath}
\usepackage{enumitem}
\usepackage{float}

\title{\texttt{THIVLVC}: Retrieval Augmented Dependency Parsing for Latin}

\name{Luc Pommeret$^{1}$, Thibault Wagret$^{2}$, Jules Deret}

\address{$^{1}$Université Paris-Saclay, CNRS, LISN, 91400, Orsay, France \\
         $^{2}$École Normale Supérieure de Lyon, HISOMA, 69007, Lyon, France \\
         pommeret@lisn.fr, thibault.wagret@ens-lyon.fr, deret.jules@gmail.com}

\abstract{
We describe \texttt{THIVLVC}, a two-stage system for the EvaLatin 2026 Dependency Parsing task. Given a Latin sentence, we retrieve structurally similar entries from the CIRCSE treebank using sentence length and POS $n$-gram similarity, then prompt a large language model to refine the baseline parse from \texttt{UDPipe} using the retrieved examples and UD annotation guidelines. We submit two configurations: one without retrieval and one with retrieval (RAG). On poetry (Seneca), \texttt{THIVLVC} improves CLAS by +17 points over the UDPipe baseline; on prose (Thomas Aquinas), the gain is +1.5 CLAS. A double-blind error analysis of 300 divergences between our system and the gold standard reveals that, among unanimous annotator decisions, 53.3\% favour \texttt{THIVLVC}, showing annotation inconsistencies both within and across treebanks.
\\ \newline }

\begin{document}

\maketitleabstract

\section{Introduction}

The EvaLatin 2026 Dependency Task \cite{iurescia-etal-2026-overview} invites participants to parse Latin texts for two genres: Classical poetry with Seneca, and philosophical prose of Thomas Aquinas.

Previous systems have relied on supervised neural models trained on existing treebanks. Such models learn whatever patterns the training data contains, including, inevitably, annotation choices that predate current guidelines. We explore a complementary approach: explicit injection of UD rules into a Large Language Model, allowing it to refine the output of a traditional parser, combined with a retrieval component. The method is simple. Whether it generalizes beyond Latin remains to be seen.

\section{Description of the System}

Our system is a two-stage pipeline\footnote{Code available at \\\url{https://github.com/l-pommeret/THIVLVC}}: (1)~retrieval of structurally similar sentences from CIRCSE, and (2)~generation, where an LLM refines a baseline parse using the retrieved examples and UD guidelines.

\begin{figure}[t]
\centering
\includegraphics[width=0.6\columnwidth]{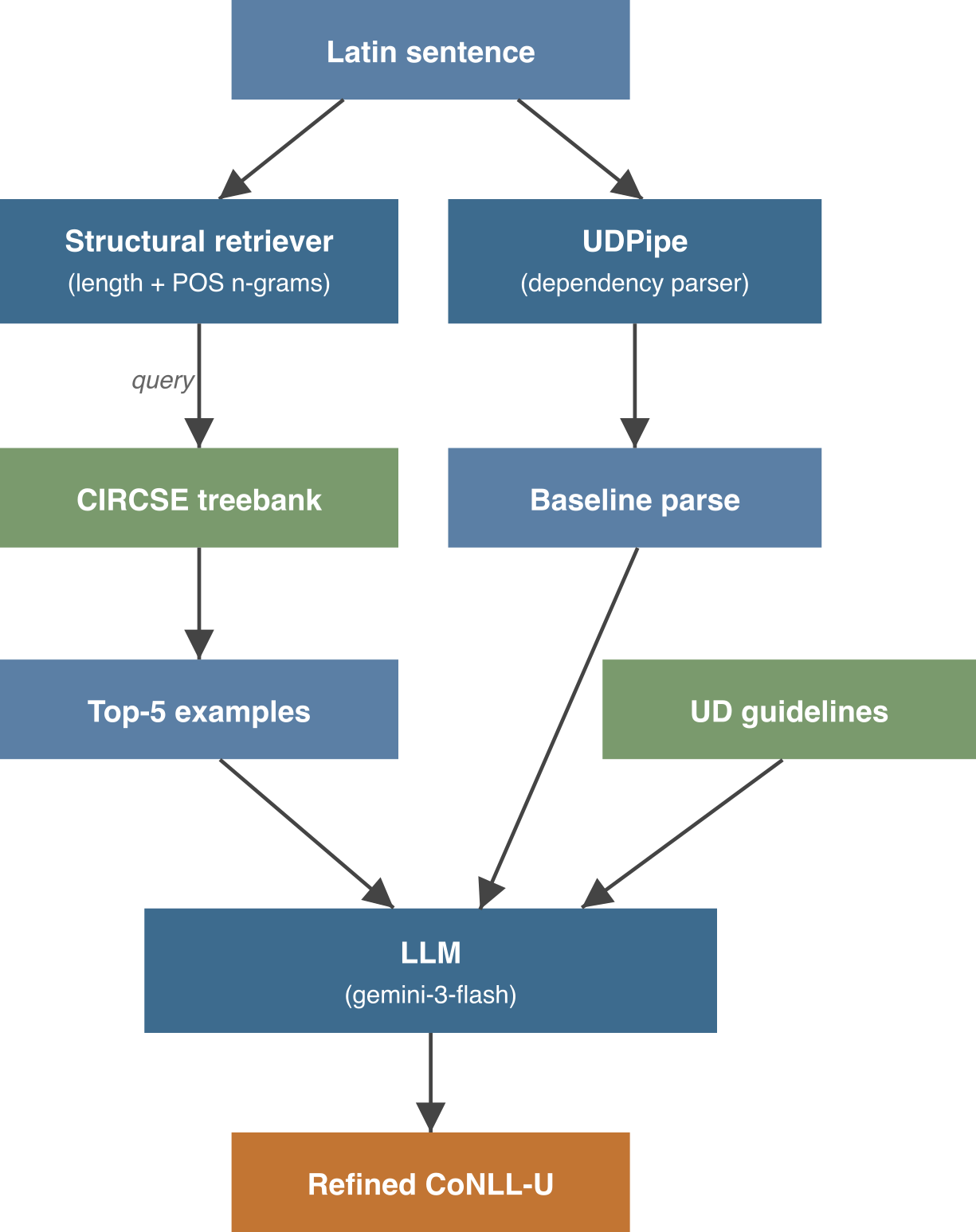}
\vspace{-1ex}
\caption{Overview of the \texttt{THIVLVC} pipeline. The input sentence is processed in parallel by the structural retriever (which selects similar examples from CIRCSE) and by UDPipe (which produces a baseline dependency parse). Both outputs, together with the UD guidelines, are passed to the LLM for refinement.}
\label{fig:system}
\vspace{-1ex}
\end{figure}

\paragraph{Information Retrieval.}
Given an input sentence, we retrieve the $k=5$ most similar sentences from the training set of the CIRCSE treebank (762~sentences) using the \emph{structural} retriever. This retriever is based on sentence length and POS $n$-grams provided in the input data. The similarity function is a weighted combination of normalized length difference and Jaccard similarities over POS bigrams and trigrams (Equation~\ref{eq:struct}).

\paragraph{Generation.}
The retrieved examples, together with the official UD annotation guidelines and the baseline parse from UDPipe, are passed to \texttt{gemini-3-flash} \citep{gemini3flash2026}. The LLM is prompted to act as a ``Latin Chief Annotator'': it compares the baseline parse with the retrieved examples and the guidelines, then outputs a refined CoNLL-U block with corrected HEAD and DEPREL columns. The full prompt is given in Appendix~\ref{sec:prompt}.

We submit two configurations: \texttt{THIVLVC\_1} (LLM + UDPipe + UD guidelines, without retrieval) and \texttt{THIVLVC\_2} (same, but with RAG on CIRCSE). Figure~\ref{fig:system} gives an overview of the pipeline.

\section{Evaluation Protocol}

\paragraph{Retrieval strategies.}
We compared three retrieval strategies. In each case, $q$ denotes the query sentence and $s$ a candidate from the knowledge base.

\begin{enumerate}[leftmargin=*,nosep]

\item TF-IDF (Baseline).
Cosine similarity where $\mathbf{v}_q$ and $\mathbf{v}_s$ are the TF-IDF vectors of word forms for $q$ and $s$:
\begin{equation*}
\mathrm{sim}_{\text{tfidf}}(q, s) = \frac{\mathbf{v}_q \cdot \mathbf{v}_s}{\|\mathbf{v}_q\| \, \|\mathbf{v}_s\|}
\end{equation*}

\item Structural (Length + POS $n$-grams).
A weighted combination of sentence length similarity and POS $n$-gram Jaccard overlap:
\begin{equation}
\label{eq:struct}
\mathrm{sim}_{\text{struct}}(q, s) = 0.33 \, f_{\text{len}}
  + 0.33 \, f_{\text{bi}} + 0.34 \, f_{\text{tri}}
\end{equation}
where $f_{\text{len}} = 1 - \tfrac{|\,|q| - |s|\,|}{\max(|q|, |s|)}$ is the normalized length similarity ($|q|$ and $|s|$ denote sentence lengths in tokens), $f_{\text{bi}} = J(\text{bigrams}(\text{POS}_q),\, \text{bigrams}(\text{POS}_s))$ is the Jaccard coefficient $J(A, B) = |A \cap B| / |A \cup B|$ over POS bigrams, and $f_{\text{tri}}$ is defined analogously for trigrams.

\item Morphological.
Cosine similarity over TF-IDF vectors of concatenated POS and morphological features (\texttt{POS|FEATS} per token).

\end{enumerate}

\paragraph{Retrieval metrics.}
Let $Q = \{q_1, \ldots, q_M\}$ be the test set. For each query $q_i$, we retrieve $k=5$ examples $s_{i,1}, \ldots, s_{i,k}$ from the knowledge base. We evaluate retrieval quality with two metrics:

\textbf{Length Difference}: the average absolute difference in sentence length (in tokens):
\begin{equation*}
\text{LenDiff} = \frac{1}{Mk} \sum_{i=1}^{M} \sum_{j=1}^{k} \bigl| |q_i| - |s_{i,j}| \bigr|
\end{equation*}
A small difference ($< 2$ tokens) indicates similar syntactic complexity.

\textbf{POS Overlap}: the average Jaccard coefficient $J$ (as defined above) over POS tag sets. For a sentence $s$, let $P(s) = \text{unique}(\text{POS}(s))$:
\begin{equation*}
\text{POSOverlap} = \frac{1}{Mk} \sum_{i=1}^{M} \sum_{j=1}^{k} J\bigl(P(q_i),\, P(s_{i,j})\bigr)
\end{equation*}
Higher is better ($[0, 1]$).

\paragraph{Generation.}
We tested three recent LLMs of varying scale and provider:
\texttt{gemini-3-flash}, \texttt{claude-4.5-sonnet}, and \texttt{qwen3-72B}.

\paragraph{Benchmarks.}
We use the CIRCSE treebank for retrieval evaluation and the EvaLatin~2026 test set for generation evaluation. The official metrics are CLAS and LAS (F1), reported both with and without relation subtypes.

\section{Results and Analysis}

\paragraph{IR results.} Table~\ref{tab:rag-comparison} shows that the structural strategy strongly outperforms TF-IDF and morphological retrieval on length difference ($< 1.2$ tokens on average vs.\ $> 11$), while maintaining competitive POS overlap. This confirms that sentence length and POS $n$-grams are sufficient features for retrieving structurally similar examples.

\paragraph{System results.} Table~\ref{tab:thivlvc} compares our two configurations with the UDPipe baseline \cite{straka-strakova-2020-udpipe}. On poetry, \texttt{THIVLVC\_2} improves CLAS by +17 points over UDPipe (with subtypes). On prose, the gain is more modest (+1.5 CLAS), as UDPipe already performs well on this genre. The RAG component (\texttt{THIVLVC\_2} vs.\ \texttt{THIVLVC\_1}) brings a consistent improvement, especially on prose (+6.9 CLAS with subtypes).

\begin{table}[t]
\centering
\setlength{\tabcolsep}{3pt}
\scriptsize
\begin{tabular}{llrr}
\toprule
\textbf{Dataset} & \textbf{Strategy} & \textbf{Length Diff} & \textbf{POS Overlap} \\
\midrule
\multirow{3}{*}{Prose}
  & TF-IDF        & $13.24 \pm 12.09$ & $0.421 \pm 0.150$ \\
  & Morphological & $12.86 \pm 14.78$ & $\mathbf{0.519} \pm 0.138$ \\
  & Structural    & $\mathbf{0.76} \pm 1.07$ & $0.512 \pm 0.142$ \\
\midrule
\multirow{3}{*}{Poetry}
  & TF-IDF        & $11.60 \pm 15.95$ & $0.454 \pm 0.177$ \\
  & Morphological & $14.51 \pm 20.63$ & $0.556 \pm 0.163$ \\
  & Structural    & $\mathbf{1.13} \pm 3.02$ & $\mathbf{0.601} \pm 0.196$ \\
\midrule
\multirow{3}{*}{Combined}
  & TF-IDF        & $12.26 \pm 14.52$ & $0.441 \pm 0.167$ \\
  & Morphological & $13.88 \pm 18.56$ & $0.541 \pm 0.154$ \\
  & Structural    & $\mathbf{0.98} \pm 2.44$ & $\mathbf{0.565} \pm 0.177$ \\
\midrule
\multirow{3}{*}{CIRCSE}
  & TF-IDF        & $11.63 \pm 14.34$ & $0.471 \pm 0.185$ \\
  & Morphological & $15.55 \pm 20.49$ & $0.568 \pm 0.175$ \\
  & Structural    & $\mathbf{1.06} \pm 1.73$ & $\mathbf{0.611} \pm 0.198$ \\
\bottomrule
\end{tabular}
\caption{Retrieval strategies comparison. Knowledge base: CIRCSE train (762~sentences). $k=5$ retrievals per query.}
\label{tab:rag-comparison}
\end{table}

\begin{table}[ht]
\centering
\setlength{\tabcolsep}{3pt}
\scriptsize
\begin{tabular}{llcccc}
\toprule
& & \multicolumn{2}{c}{\textbf{With subtypes}} & \multicolumn{2}{c}{\textbf{No subtypes}} \\
\cmidrule(lr){3-4} \cmidrule(lr){5-6}
\textbf{System} & \textbf{Genre} & \textbf{CLAS} & \textbf{LAS} & \textbf{CLAS} & \textbf{LAS} \\
\midrule
\multirow{2}{*}{\texttt{UDPipe}} & Poetry & 56.94 & 57.22 & 57.24 & 59.74 \\
& Prose & 79.41 & 82.17 & 83.07 & 85.21 \\
\midrule
\multirow{2}{*}{\texttt{THIVLVC\_1}} & Poetry & 72.71 & 70.36 & 76.00 & 76.97 \\
& Prose & 74.04 & 75.72 & 81.52 & 82.78 \\
\midrule
\multirow{2}{*}{\texttt{THIVLVC\_2}} & Poetry & \textbf{74.03} & \textbf{72.88} & \textbf{76.08} & \textbf{77.60} \\
& Prose & \textbf{80.92} & \textbf{83.26} & \textbf{86.60} & \textbf{87.93} \\
\bottomrule
\end{tabular}
\caption{\texttt{THIVLVC} system comparison. \texttt{THIVLVC\_1} = LLM + UDPipe + UD Guidelines. \texttt{THIVLVC\_2} = \texttt{THIVLVC\_1} + RAG on CIRCSE.}
\label{tab:thivlvc}
\end{table}

\section{Error Analysis}

Not all divergences from the gold standard are errors. To better understand our system's behaviour, we conducted a qualitative analysis of cases where predictions differed from reference annotations.

\paragraph{Annotation protocol.} We designed a double-blind annotation comparing Gold and \texttt{THIVLVC} outputs (see the interface in Figure~\ref{fig:interface}). Annotators were presented with divergent annotations without knowing which came from which system, and chose among five categories: ``Gold is better'', ``System is better'', ``both wrong'', ``undecidable'', and ``don't know''.

\begin{table}[ht]
\centering
\setlength{\tabcolsep}{3pt}
\scriptsize
\begin{tabular}{lrrrr}
\toprule
\textbf{Verdict} & \multicolumn{2}{c}{\textbf{Ann.\ 1}} & \multicolumn{2}{c}{\textbf{Ann.\ 2}} \\
\cmidrule(lr){2-3} \cmidrule(lr){4-5}
& $n$ & \% & $n$ & \% \\
\midrule
Gold (human) & 137 & 45.7 & 110 & 36.7 \\
\texttt{THIVLVC} (system)  & 126 & 42.0 & 143 & 47.7 \\
Both wrong   &   7 &  2.3 &  11 &  3.7 \\
Undecidable  &  11 &  3.7 &  21 &  7.0 \\
Don't know   &  19 &  6.3 &  15 &  5.0 \\
\bottomrule
\end{tabular}
\caption{Blind evaluation verdicts per annotator (300~items).}
\label{tab:annotator-dist}
\end{table}

Table~\ref{tab:annotator-dist} shows that out of 300 divergences, annotators unanimously agreed on 167~cases, of which 89 (53.3\%) favour \texttt{THIVLVC} and 78 (46.7\%) the gold standard (Table~\ref{tab:consensus}). Tables~\ref{tab:taxonomy} and~\ref{tab:confusions} in the Appendix break these down by error type and most frequent label confusions. Drawing on UD guidelines and recent work on Latin treebank harmonization \citep{gamba-zeman-2023-universalising}, we organise the discussion into five categories.

\section{Taxonomy of Disagreement}

\paragraph{Contradictions between CIRCSE and EvaLatin.}
According to Table~\ref{tab:confusions}, errors of the type \texttt{advmod:lmod} instead of \texttt{advmod} account for 7 out of 37 (18\%) of the main \texttt{THIVLVC} errors.
The adverb \textit{unde} illustrates a case of legitimate annotation divergence between the CIRCSE corpus and the EvaLatin~2026 corpus. In CIRCSE,\footnote{Sentence \texttt{Latin\_Tacitus\_Ger\_prose-163}: \textit{unde annum quoque ipsum non in totidem digerunt species hiems et uer} [\ldots] (``Hence they do not divide the year itself into the same number of seasons: winter, spring~[\ldots]'').} \textit{unde} is annotated \texttt{advmod:lmod}, signalling a spatial reference in the underlying description. In EvaLatin~2026, by contrast, \textit{unde} is annotated as a simple \texttt{advmod}, without the subtype \texttt{:lmod}.\footnote{E.g.\ sentence~s142: \textit{unde sacra doctrina maxime dicitur sapientia} (``Hence sacred doctrine is called wisdom in the highest sense'').}
Both analyses are linguistically defensible: \textit{unde} may simultaneously carry a spatial origin reading (justifying \texttt{:lmod}) and function as a logical connector, and neither interpretation excludes the other. The two corpora simply reflect different (but individually coherent) annotation conventions regarding the scope of the \texttt{:lmod} subtype.

Our system reproduces the \texttt{advmod:lmod} pattern learned from the CIRCSE training data. When evaluated against EvaLatin~2026, this behaviour is penalized, although it reflects a valid annotation choice. This type of divergence suggests that evaluation metrics should ideally accept both annotations as correct when two conventions are independently defensible, rather than treating the test set annotation as the sole ground truth.

This case illustrates a broader methodological issue in dependency parsing: an annotation shift between training and evaluation datasets. Similar inconsistencies across treebanks have been documented in previous work on UD harmonization, especially among subtypes.\footnote{``The most widespread issues are the \texttt{tmod} and \texttt{lmod} relation subtypes, as well as comparative clauses.'' \cite{gamba-zeman-2023-universalising}.} When a training corpus uses a subtype that the evaluation corpus omits, parsers may be penalized for faithfully reproducing the annotation patterns present in their training data.

\paragraph{Internal gold inconsistency: \texttt{obl} vs \texttt{obl:arg}.}
According to Table~\ref{tab:confusions}, errors of the type \texttt{obl} instead of \texttt{obl:arg} and \texttt{obl} instead of \texttt{obl:lmod} account for 12 out of 26 (46.2\%) of the main EvaLatin~2026 errors.
Some discrepancies between \texttt{THIVLVC} and EvaLatin~2026 arise from internal inconsistencies within EvaLatin~2026 itself, particularly in the use of the \texttt{obl:arg} subtype (the same goes for \texttt{obl:lmod}). EvaLatin~2026 contains divergent annotations for the verb \textit{pertineo} and its prepositional complement introduced by \textit{ad}.\footnote{In sentence~S2, \textit{propositum nostrae intentionis in hoc opere est, ea quae ad christianam religionem pertinent} [\ldots], \textit{religionem} is annotated \texttt{obl:arg} of \textit{pertinent}. However, in sentence~s147, \textit{cum iudicium ad sapientem pertineat} [\ldots], \textit{sapientem} governed by the same verb with the same preposition, is annotated as bare \texttt{obl}.} The two examples receive different annotations despite their identical syntactic configuration. UD guidelines recommend \texttt{obl:arg} for oblique arguments selected by a predicate.\footnote{\url{https://universaldependencies.org/u/dep/obl-arg.html}}

\texttt{THIVLVC}, by contrast, annotates \texttt{obl:arg} in both cases, consistently with its training on CIRCSE and with the UD guidelines. The discrepancy therefore appears to result from variation in annotation practice rather than from a parsing error.

Such intra-corpus inconsistencies are distinct from the inter-corpus divergences discussed above. They suggest that EvaLatin~2026 contains a degree of internal annotation noise. Similar annotation fluctuations within the same dataset have been documented in previous work on UD harmonization \citep{gamba-zeman-2023-universalising}. The distinction between arguments and oblique modifiers is widely recognized as difficult to apply consistently in dependency annotation \cite{de-marneffe-etal-2014-universal}.

\paragraph{Clear-cut gold error: \texttt{mark} for \texttt{case}.}
Sentence~s127 of EvaLatin~2026 illustrates a likely human annotation error.\footnote{In \textit{sed haec doctrina supponit principia sua aliunde, ut ex dictis patet}, the word \textit{ex} is annotated as \texttt{mark}. The same preposition is correctly annotated as \texttt{case} in sentence~s47: \textit{omnis enim scientia procedit ex principiis per se notis}.} The relation \texttt{mark} is reserved in the UD guidelines for subordinating conjunctions and clause-introducing function words \citep{de-marneffe-etal-2014-universal}, whereas prepositions governing noun phrases are annotated as \texttt{case}. \texttt{THIVLVC} correctly annotates \texttt{case} in both instances, demonstrating consistency with the UD guidelines.

Similar misattributions of functional relations have been observed in Latin treebanks during harmonization and conversion processes. For instance, \cite{gamba-zeman-2023-universalising} report that nominal obliques were sometimes misannotated as \texttt{advmod} in PROIEL, while \cite{cecchini-etal-2018-enhancing} show that the conversion of the \textit{Index Thomisticus} from Prague Dependency style to UD introduced systematic errors in prepositional functional relations.

\paragraph{Undecidable ambiguity: \textit{hic} (\texttt{det} or \texttt{advmod:lmod}?).}
Not all divergences between the parser and EvaLatin~2026 can be attributed to annotation error. Some reflect genuine linguistic ambiguity that the UD guidelines do not fully resolve.\footnote{In the CIRCSE sentence \textit{hic rapax torrens cadit} [\ldots] (SenPhoen-P-17-8), the form \textit{hic} can be analyzed either as an anaphoric determiner modifying \textit{torrens} (\texttt{det}: ``this violent torrent here falls'') or as a locative adverb attached to \textit{cadit} (\texttt{advmod:lmod}: ``here a violent torrent falls''). Both readings are syntactically and philologically defensible.}
Such cases illustrate a structural limitation of single-analysis treebank annotation. Because a gold standard records only one syntactic analysis, alternative interpretations are necessarily excluded, and parsers are penalized when they select a different but linguistically plausible structure. Cases of this kind have long been identified as a general challenge for treebank evaluation \cite{gamba-zeman-2023-universalising}.

\paragraph{Errors linked with punctuation.}
Some groups of words presented as sentences in the corpus appear to contain several sentences.\footnote{For example, \textit{omitte poenae languidas longae moras mortemque totam admitte quid segnis traho quod uiuo} (SenPhoen-P-17-22, Loeb: ``Away with the slow delays of thy long-due punishment; receive death wholly. Why do I sluggishly drag on this life?''). Both the Latin text and the translation distinguish two sentences, separated by a full stop, also in the Teubner edition.} If some editions read two sentences, the dependency between the root (\textit{omitte}) and the head of the second clause (\textit{traho}) could be analyzed as \texttt{parataxis} rather than \texttt{conj}. However, editorial punctuation reflects interpretive choices rather than established facts about the text, and the corpus annotation is not necessarily wrong for diverging from a given printed edition. EvaLatin~2026, however, reads \texttt{conj(omitte, traho)}, while \texttt{THIVLVC} has \texttt{parataxis(omitte, traho)}.

In another sentence, both EvaLatin~2026 and \texttt{THIVLVC} appear to make the same interpretation.\footnote{In \textit{flammas potius et uastum aggerem compone in altos ipse me immittam rogos} [\ldots] (SenPhoen-P-17-61), both annotate \texttt{conj}(\textit{compone}, \textit{immittam}). The Loeb edition, however, distinguishes the two clauses with a semicolon, suggesting \texttt{parataxis}(\textit{compone}, \textit{immittam}).} The lack of punctuation in the corpus can mislead both human annotators and automatic systems. A solution to this problem could be to better segment sentences in the corpus.

\paragraph{Adjectival misclassification: \texttt{amod} for \texttt{acl}.} According to Table \ref{tab:confusions}, errors of the type \texttt{amod} instead of \texttt{acl} account for 8 out of the 37 most frequent gold-correct cases (21.6\%), making it one of the most frequent \texttt{THIVLVC} error types. In these instances, the system annotates past participles such as \textit{subiecti}, \textit{scissa} or \textit{lapsi} as \texttt{amod} rather than \texttt{acl}, which is the annotation expected by the UD guidelines for participial constructions retaining verbal properties. The confusion is not entirely surprising from a linguistic standpoint: in traditional grammar, the participle is conventionally described as the adjectival form of the verb. The case of \textit{notus} illustrates this ambiguity well: originally a participle of \textit{nosco}, it has undergone a degree of lexicalisation and functions in many contexts as a plain adjective ("well-known"), making either analysis defensible. Nevertheless, the systematic nature of the misinterpretation (affecting multiple participles across different sentences) remains difficult to account for. The CIRCSE training data consistently distinguishes \texttt{acl} from \texttt{amod} in comparable configurations, and the UD guidelines passed to \texttt{THIVLVC} explicitly define \texttt{acl} as the appropriate relation for participial modifiers of nominals.

\section{Limitations}
Our approach has several limitations. First, the selection of the LLM was based on informal manual comparison rather than a systematic ablation study. We tested \texttt{gemini-3-flash}, \texttt{claude-4.5-sonnet}, and \texttt{qwen3-72B} on a small set of sentences and selected \texttt{gemini-3-flash} on the basis of output quality and cost, but we did not conduct a controlled evaluation across models. This limits the reproducibility and generalizability of our LLM choice.
Second, the system relies on a single knowledge base (CIRCSE, 762~sentences). The retriever cannot return useful examples for syntactic constructions absent from this small corpus, which may explain the more modest gains on prose, where UDPipe already performs well.
Third, LLM-based parsing is inherently non-deterministic: the same prompt can yield different outputs across runs. We did not measure this variance. A related concern is cost: each sentence requires an API call, making the approach substantially more expensive than a fine-tuned neural parser.
Finally, our error analysis, while informative, is limited to 300~items annotated by two annotators with fair inter-annotator agreement ($\kappa = 0.49$ on the binary Gold/\texttt{THIVLVC} decision). A larger annotation campaign would be needed to draw firmer conclusions.

\section{Conclusion}
We presented \texttt{THIVLVC}, a retrieval-augmented LLM system for Latin dependency parsing that combines a structural retriever, UD annotation guidelines, and a baseline parse to refine syntactic analysis. The system achieves substantial improvements over UDPipe on poetry and competitive results on prose.
Our error analysis highlights a finding that goes beyond system performance: a significant proportion of divergences between parser output and gold annotations stem from annotation inconsistencies, both across corpora (CIRCSE vs.\ EvaLatin~2026) and within a single dataset. These results underscore the need for continued harmonization efforts in Latin treebanks, in line with previous work \citep{gamba-zeman-2023-universalising}.
More broadly, our analysis suggests that evaluation protocols for dependency parsing could benefit from accepting multiple valid annotations in cases where legitimate annotation conventions diverge or where genuine linguistic ambiguity makes a single gold standard inadequate.
In future work, we plan to conduct a systematic comparison of LLMs for this task, expand the retrieval knowledge base to multiple treebanks, and investigate whether fine-tuning a smaller model on LLM-generated corrections could reduce inference cost while preserving quality.

\section{Acknowledgements}
We thank the EvaLatin 2026 organizers for making the shared task data available, and the CIRCSE Research Centre for the treebank used as knowledge base. This work was carried out at LISN (CNRS, Université Paris-Saclay) and HISOMA (ENS of Lyon).

\vskip30pt

\section{Bibliographical References}

\bibliographystyle{lrec2026-natbib}
\bibliography{evalatin2026}

\clearpage

\appendix

\section{Prompt}
\label{sec:prompt}

The LLM receives the following prompt (the variables are filled at runtime):

\small
\begin{verbatim}
You are the Latin Chief Annotator.
Your goal: Refine and improve the syntax
(HEAD/DEPREL) of the Input Sentence using
best practices.

=== OFFICIAL ANNOTATION GUIDELINES ===
{guidelines_text}
======================================

Here are 5 SIMILAR examples from the
training data.
{example_str}

--- BASELINE (from automatic parser) ---
{latinpipe_context}

--- Input Sentence ---
{conllu_str}

Task:
1. Compare the baseline parse with the
   examples and guidelines
2. Identify any improvements needed in
   HEAD/DEPREL
3. Output your refined version
4. If uncertain, add a comment line
   # needs_council = true

Output ONLY the CoNLL-U block for the
Input Sentence (not the baseline).
\end{verbatim}
\normalsize

\section{Annotation Interface}
\label{sec:interface}

\begin{figure}[H]
\centering
\includegraphics[width=\columnwidth]{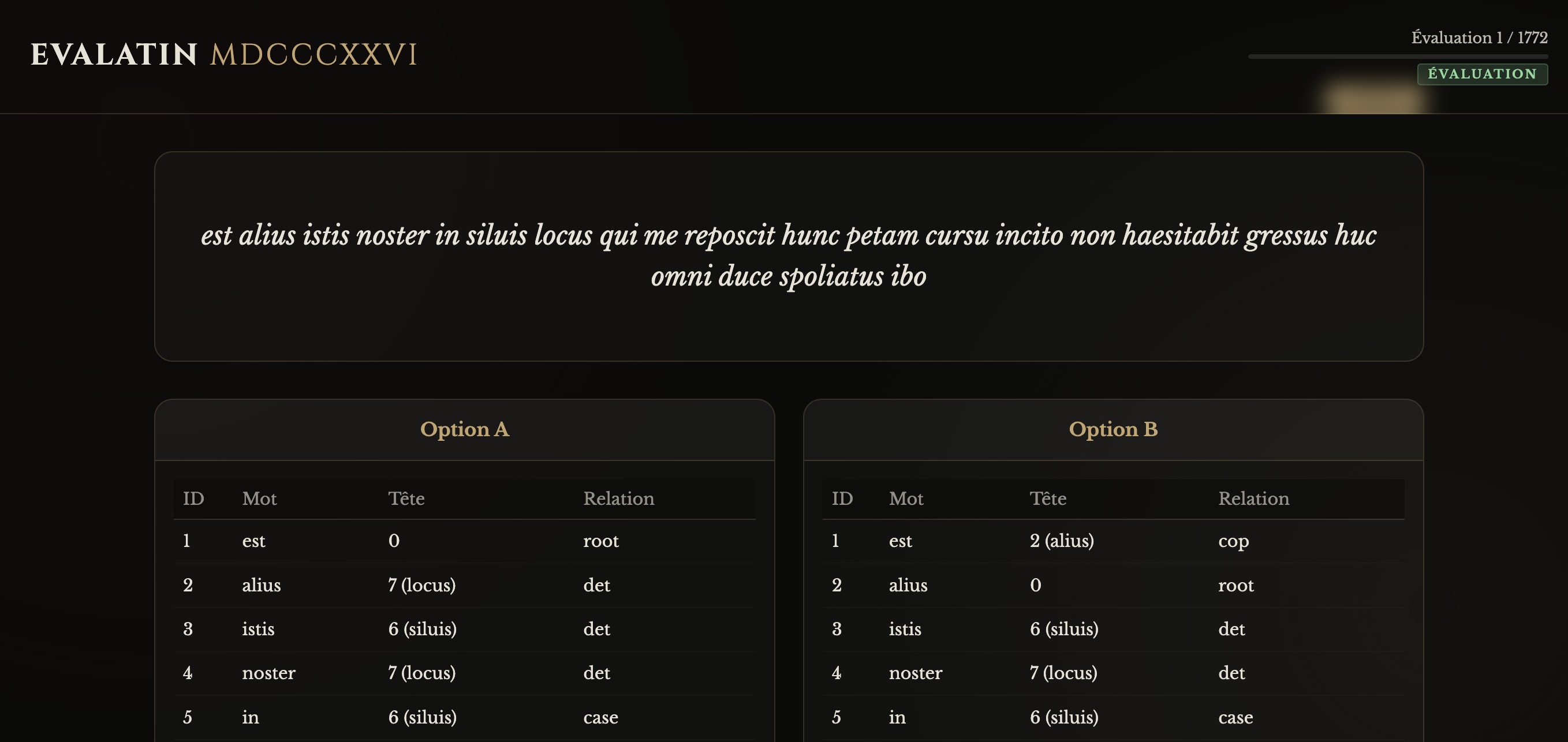}
\caption{Annotation interface: overview. The Latin sentence is displayed at the top; two anonymized parse options (A and B) are shown side by side with their ID, word form, head, and relation columns.}
\label{fig:interface}
\end{figure}

\begin{figure}[H]
\centering
\includegraphics[width=\columnwidth]{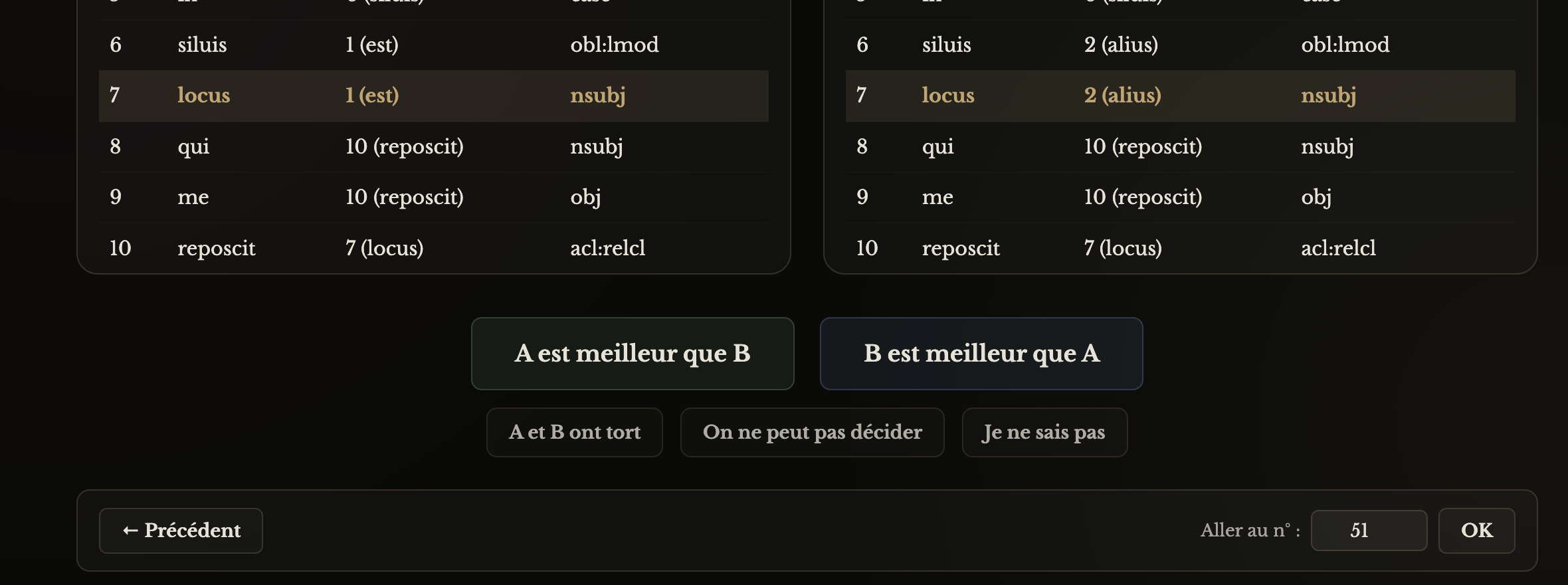}
\caption{Annotation interface: verdict buttons. Divergent rows are highlighted. Annotators choose among five categories.}
\label{fig:interface-buttons}
\end{figure}

\section{Detailed Annotation Results}

We provide detailed statistics to support the comparison between \texttt{THIVLVC} and the gold standard.

Table~\ref{tab:consensus} shows that, overall, \texttt{THIVLVC} is more often correct than the gold in head-to-head comparisons, at least on items where both annotators reach a consensus.

\begin{table}[H]
\centering
\caption{Unanimous consensus: items where both annotators agree (173/300, 57.7\%).}
\label{tab:consensus}
\begin{tabular}{lrr}
\toprule
\textbf{Verdict} & $n$ & \% \\
\midrule
\texttt{THIVLVC} better than Gold & 89 & 53.3 \\
Gold better than \texttt{THIVLVC} & 78 & 46.7 \\
\midrule
\textit{Decided} & \textit{167} & \\
\midrule
Undecidable & 3 & 1.7 \\
Don't know  & 3 & 1.7 \\
\midrule
\textbf{Total} & \textbf{173} & \\
\bottomrule
\end{tabular}
\end{table}

When examining the different error types in Table~\ref{tab:taxonomy}, we observe that the most frequent errors are distributed across head attachment, relation labelling, and subtype confusion.

\begin{table}[H]
\centering
\caption{Taxonomy of errors by source (Gold vs.\ \texttt{THIVLVC}), based on 167 unanimous decided cases.}
\label{tab:taxonomy}
\small
\begin{tabular}{lrrrr}
\toprule
\textbf{Error type} & \multicolumn{2}{c}{\textbf{Gold err.}} & \multicolumn{2}{c}{\textbf{\texttt{THIVLVC} err.}} \\
\cmidrule(lr){2-3} \cmidrule(lr){4-5}
& $n$ & \% & $n$ & \% \\
\midrule
Wrong head + relation & 26 & 29.2 & 14 & 17.9 \\
Subtype confusion     & 23 & 25.8 & 29 & 37.2 \\
Wrong head only       & 20 & 22.5 &  8 & 10.3 \\
Wrong relation only   & 17 & 19.1 & 24 & 30.8 \\
Head + subtype        &  3 &  3.4 &  3 &  3.8 \\
\midrule
\textbf{Total} & \textbf{89} & & \textbf{78} & \\
\bottomrule
\end{tabular}
\end{table}

The label confusions are also informative: Table~\ref{tab:confusions} shows that the most frequent gold errors involve \texttt{obl}, while the most frequent \texttt{THIVLVC} errors involve \texttt{amod} and \texttt{advcl}.

\begin{table}[H]
\centering
\caption{Most frequent label confusions by error source.}
\label{tab:confusions}
\resizebox{\columnwidth}{!}{%
\begin{tabular}{llr|llr}
\toprule
\multicolumn{3}{c|}{\textbf{Gold errors (\texttt{THIVLVC} correct)}} & \multicolumn{3}{c}{\textbf{\texttt{THIVLVC} errors (Gold correct)}} \\
\cmidrule(lr){1-3} \cmidrule(lr){4-6}
Wrong & Correct & $n$ & Wrong & Correct & $n$ \\
\midrule
obl       & obl:arg  & 6 & amod        & acl          & 8 \\
obl       & obl:lmod & 6 & advcl       & advcl:cmp    & 8 \\
amod      & root     & 4 & advmod      & advmod:tmod  & 7 \\
obl       & nummod   & 4 & advmod:lmod & advmod       & 7 \\
conj      & parataxis& 3 & advmod      & discourse    & 4 \\
obl:agent & obl:arg  & 3 & conj        & conj:expl    & 3 \\
\bottomrule
\end{tabular}%
}
\end{table}

Regarding genre, Table~\ref{tab:genre} shows that the consensus is fairly evenly distributed between poetry and prose.

\begin{table}[H]
\centering
\caption{Consensus results by genre (167 decided cases).}
\label{tab:genre}
\begin{tabular}{lrrrr}
\toprule
\textbf{Genre} & \textbf{Gold} & \textbf{\texttt{THIVLVC}} & \textbf{Total} & \textbf{\texttt{THIVLVC} \%} \\
\midrule
Poetry & 49 & 50 & 99 & 50.5 \\
Prose  & 29 & 39 & 68 & 57.4 \\
\midrule
\textbf{All} & \textbf{78} & \textbf{89} & \textbf{167} & \textbf{53.3} \\
\bottomrule
\end{tabular}
\end{table}

Table~\ref{tab:iaa} shows that inter-annotator agreement is reasonably solid, with a Cohen's $\kappa$ of 0.493 on head-to-head Gold vs.\ \texttt{THIVLVC} items.

\begin{table}[H]
\centering
\caption{Inter-annotator agreement on 300 blind evaluation items. $\kappa$~is Cohen's kappa.}
\label{tab:iaa}
\resizebox{\columnwidth}{!}{%
\begin{tabular}{lcc}
\toprule
\textbf{Metric} & \textbf{All categories} & \textbf{Gold/\texttt{THIVLVC} only} \\
\midrule
Items evaluated            & 300   & 224   \\
Observed agreement ($p_o$) & 0.577 & 0.746 \\
Expected agreement ($p_e$) & 0.374 & 0.499 \\
Cohen's $\kappa$           & 0.324 & 0.493 \\
\bottomrule
\end{tabular}%
}
\end{table}

Table~\ref{tab:confusion} presents the full confusion matrix between annotators, showing that most disagreements arise when one annotator selects Gold and the other selects \texttt{THIVLVC}.

\begin{table}[H]
\centering
\caption{Confusion matrix between annotators (300~items). Rows = Ann.~1, Columns = Ann.~2.}
\label{tab:confusion}
\resizebox{\columnwidth}{!}{%
\begin{tabular}{lrrrrr|r}
\toprule
& \textbf{Gold} & \textbf{\texttt{THIVLVC}} & \textbf{Both} & \textbf{Undec.} & \textbf{DK} & \textbf{Total} \\
\midrule
\textbf{Gold}   & \textbf{78} & 37 & 6 &  8 & 8 & 137 \\
\textbf{\texttt{THIVLVC}}     & 20 & \textbf{89} & 3 & 10 & 4 & 126 \\
\textbf{Both}   &  3 &  4 & 0 &  0 & 0 &   7 \\
\textbf{Undec.} &  3 &  5 & 0 & \textbf{3} & 0 &  11 \\
\textbf{DK}     &  6 &  8 & 2 &  0 & \textbf{3} &  19 \\
\midrule
\textbf{Total}  & 110 & 143 & 11 & 21 & 15 & 300 \\
\bottomrule
\end{tabular}%
}
\end{table}

\end{document}